# On the definition of Shape Parts: a Dominant Sets Approach

Foteini Fotopoulou[+] and George Economou[*]

[+] Department of Computer Engineering and Informatics, University of Patras, 26500, Greece
Email: fotopoulou@ceid.upatras.gr

[*] Department of Physics, University of Patras, 26500, Greece
Email: economou@physics.upatras.gr

## Abstract
In the present paper a novel graph-based approach to the shape decomposition problem is addressed. The shape is appropriately transformed into a visibility graph enriched with local neighborhood information. A two-step diffusion process is then applied to the visibility graph that efficiently enhances the information provided, thus leading to a more robust and meaningful graph construction. Inspired by the notion of a clique as a strict cluster definition, the dominant sets algorithm is invoked, slightly modified to comport with the specific problem of defining shape parts. The cluster cohesiveness and a node participation vector are two important outputs of the proposed graph partitioning method. Opposed to most of the existing techniques, the final number of the clusters is determined automatically, by estimating the cluster cohesiveness on a random network generation process. Experimental results on several shape databases show the effectiveness of our framework for graph based shape decomposition.

# 1. Introduction

Identifying a shape's components can be essential for object recognition, object completion, and shape matching [1], among other computer vision tasks [2]. The shape decomposition can be regarded without loss of generality as a clustering procedure, where the aim is to partition all of the boundary points into semantic groups. Nowadays many popular clustering approaches are based on the use of pairwise distances and graph techniques. These algorithms treat the problem of clustering as a graph partitioning one without making specific assumptions on the form of clusters. For our problem also and despite the existence of numerous shape decomposition methods, we will focus on those that map the points to be clustered into an appropriately constructed graph.

Evidently constructing the right graph is of great importance since it will influence to a large degree the final partition. A common binary graph that contains essential information for the shape structure and maps the visibility relations among nodes is the Visibility Graph [3], which is constructed by linking the mutually visible nodes, i.e. the shape boundary points that can "see" each other. Information provided by this graph is useful in shape modeling as points that are mutually visible are in convex position with respect to each other and is known that convexity plays a key role for shape decomposition [4].

This visibility graph is the starting point of our method. The information of this graph is further enhanced by appropriately transforming it into a diffusion graph, where the visibility between nodes is spatially extended in an efficient way through the two-step diffusion process and complies with the clique graph properties. Our objective is to model local neighborhoods between the data into the graph, in a way that facilitates the subsequent graph decomposition method. Such similarity graphs include the ε-neighborhood graph, the k-nearest neighbor graphs, etc [5, 6]. In the proposed method we capture local neighborhood of a graph [7], by restricting the connectivity of each node over a predefined area, appropriately computed for each shape.

Once the graph construction is completed, the graph partitioning algorithm follows, which is also of great importance. There exists a plethora of graph clustering algorithms [8]; as for example the technique described in [9] which uses the normalized graph Laplacian followed by an eigenanalysis procedure and a k-means clustering algorithm, or basic techniques that use the notion of graph-cuts [8]. Given a similarity matrix, a simple way to partition it into meaningful classes is to solve the mincut problem. However, in practice it does not always lead to satisfactory partitions, separating sometimes a single node from the rest of the graph. A way to bypass this problem is to use the RatioCut [10] and the normalized Ncut [11] objective functions. Relaxed versions of these problems can be solved using spectral clustering techniques. For example, in the case of two clusters, the minimizer of RatioCut is approximated by the signed version of the second eigenvector of the graph Laplacian (i.e. the fiedler vector), while for more clusters a k-means algorithm is needed. The k-means clustering is also associated to the solution of Ncut. In these methods the necessity of providing the number k of clusters arises naturally. Well known graph-theoretic algorithms that search for certain structure in the similarity graph, include the minimal spanning tree [12], or the method of exploring the graph in order to find a complete subgraph, which is called a clique[1]. Other authors [13] state that a maximal clique is a notion similar to the cluster definition. Although graph algorithms that use the minimal spanning tree or the normalized cuts are explicitly applied to weighted graphs, algorithms that try to discover the clique usually operate on binary graphs. An efficient generalization of the notion of maximal clique in the context of weighted graphs is developed by [14] which was first introduced in [15]. This method provides a measure of each cluster cohesiveness as well as of node participation to each group. The number of clusters is efficiently computed by using random network generation models, combined with the cluster cohesiveness of the dominant sets.

In this paper we address the shape decomposition issue as follows: each shape is represented by a visibility graph which is constructed directly from the shape, following certain rules. The visibility graph is further constrained so as to capture local neighborhood's information. However, in order to extend the visibility in an efficient way that will definitely facilitate the decomposition task, the binary graph is transformed into a weighted graph which is the output of a two-step diffusion process. After the graph construction stage, the clustering scheme takes place. In this paper the dominant sets algorithm [13,14] is applied, however appropriately modified to fit the decomposition problem. Using this method, the output clusters are computed sequentially (at decreasing order of "dominance"), putting aside the need for k-means or another extra process to identify the number of clusters. Notice that the dominant set algorithm terminates until all shape's nodes are assigned to a team. In some cases, this may be not appropriate, as some boundary points, which are usually not consecutive, should either be placed into the same cluster (the proposed modified version of the algorithm does not allow this) or should not belong to any cluster. However, we are given the possibility to find the most prominent of all groups, which may be additionally desirable in cases, for example, where we are not interested in isolating very small shape components. Therefore, in the clustering stage an extra step – which includes a graph randomization technique- is also adopted that determines the optimal number of output clusters. Finally, an optional step can be also utilized, in a post processing procedure which is based on the notion of visibility and further improves the results.

The remainder of the paper is organized as follows: In Section 2 literature overview is briefly reviewed. In Section 3 the proposed shape decomposition problem is formulated and described in detail. In Section 4 qualitative as well as quantitative results of the experiments we have conducted in several shape databases are presented. Finally, the paper concludes at Section 5.

## 2. Literature Overview

Decomposing a shape into its meaningful parts is not an easy task due to the involvement of the human's perception. This is the reason why some shape decomposition methods try to represent the shapes by some visual feature that reflects human perception [16], [17]. However, there exist some rules, borrowed from the field of cognitive theory. The most important are the minima rule [18] and the short cut rule [19]. In the former the negative curvature minima of the bounding contour (i.e. convex parts) are considered as potential points of shape separation while in the latter rule the assumption that humans divide a shape using the shortest possible cuts is present. In the following paragraph methods that use, but not limited to, the above mentioned rules are briefly described.

---

[1] In an undirected graph a clique is a subset of its vertices such that every two vertices in the subset are connected by an edge.

In [4] both convex and non-convex shape parts are used for shape decomposition. While the convex ones are easily acquired, the non-convex parts are captured by using a contour evolution procedure. Continuing, in [20] the shape is decomposed by the usage of two cuts, i.e. cut the shape in a way that the remaining one has the simplest structure or cut out the part so that this part itself takes on a simple shape. The authors of [21] propose an algorithm that computes the constrained Delaunay triangulation of a given polygonal description of a shape and chooses among the interior edges of this triangulation an optimum set of cuts by solving a corresponding combinatorial optimization problem. Moreover, in [22] a shape is decomposed into approximate convex parts using a polygon (including polygons with holes) representation. A user-tuning parameter determines the convex tolerance. Stricter convex shape decomposition is proposed in [23] where the notion of "mutex pairs" is used in an attempt to apply concavity constraints in the formalized linear programming shape decomposition problem. Due to the fact that the strict convex decomposition leads to a large number of shape parts, an approximation of convex shapes is proposed by [24]. The problem is formulated as a combinatorial optimization problem by minimizing the number of non-intersecting cuts. The convexity tolerance is specified by a user defined parameter.

Beyond the above mentioned shape partitioning methods, there exist some techniques which fall within the "graph-based" decomposition methods. These algorithms always originate from an appropriately constructed graph and use this representation in order to capture the shape parts. The method proposed by the present authors also belongs to the already mentioned category, so we consider it essential to review some of these shape decomposition techniques here. In [25] a 2D shape is represented by a binary graph followed by a graph theoretic clustering method in order to partition it into clusters that intuitively correspond to shape parts. Another interesting graph-based method, which is proposed in [9] originates from a binary visibility graph which tabulates the similarity relations between nodes, normalized in the spirit of the Laplacian formulation, but not capturing local neighborhood. An eigenanalysis follows while the parameter that defines the output number of clusters has to be extensively searched by using the k-means algorithm. A method that employs a spectral algorithm, starting with a binary matrix of mutually visible nodes, with additional boundary neighborhood restrictions is proposed in [26]. The clustering methodology is appropriately formulated as a constrained NMF problem. Finally, one more graph-based shape partition algorithm that has reviewed accurate and compact shape parts' representation is presented in [7]. This method also originates from the visibility between boundary nodes and finds shape segments by using an appropriate iterative algorithm.

## 3. The proposed method

In this section a detailed description of the introduced scheme is provided. Our method is composed of two distinct steps, i.e. the graph construction and the graph partitioning. Particularly, in the first step, the appropriate graph is constructed. For that reason, the visibility matrix is used, modified to capture the local neighborhood of the shape. Then, the visibility is spatially extended and is tabulated in the diffusion matrix. The objective of the second step is to provide an efficient graph partitioning method in order to obtain the clusters (i.e. shape parts), which is accomplished by the dominant sets algorithm. This algorithm is interchangeably used with a graph randomization technique. The aim of the latter is to automatically determine the subset of clusters that are meaningful for the shape decomposition task. A block diagram of the introduced method is depicted in Fig.1. Notice that an optional block is placed as a last step of the whole procedure, which aims to further improve (if needed) the output results.

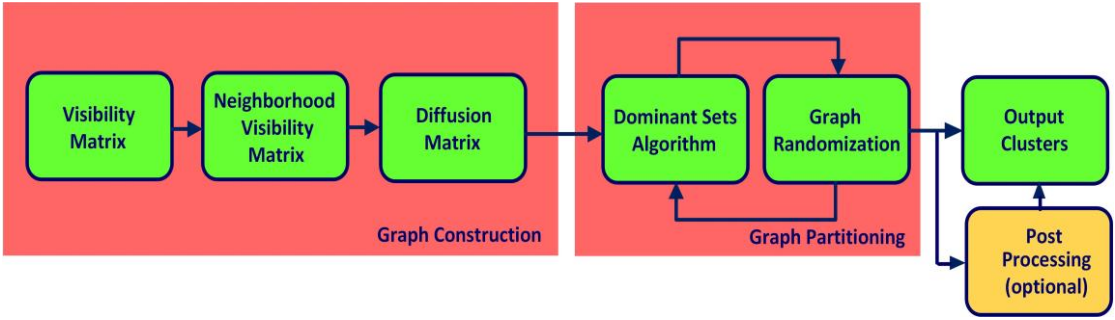

**Fig.1** A block diagram of the introduced method showing the basic steps of the procedure, i.e. the graph construction and the graph partitioning, that result in the output clusters. An optional post-processing step is also included in the end of the whole procedure.

## 3.1 The visibility matrix

In order to achieve our goal, which is decomposing a shape into its meaningful parts, the shape in the first place should be transformed into a graph. Note that the plane curve that describes the shape boundary is represented by a set of *N* position vectors:

$$r(i)=(x(i), y(i)), i=1,2,..., N. \qquad (1)$$

A shape graph is easily constructed by denoting the shape's boundary points and the lines linking these points as the graph's nodes and the graph's edges, respectively. To that end, let us consider the following undirected Visibility Graph $G_{vis}=(V, E, A)$ where V is defined as the node set denoted by Eq.1 , E is the corresponding edge set, linking these nodes and *A* is an *NxN* binary matrix, which we call the Visibility Matrix. In matrix *A* node connections are tabulated by 1'es. However, not any line connecting these nodes has physical significance for the shape representation: consider an observer placed at each node, a more natural result could be obtained by linking the nodes that fall within the observer's field of view. Therefore, the *NxN* binary, symmetric Visibility Matrix $A[a_{ij}]$ can be defined as follows:

$$a_{ij}= \begin{cases} 1 & \text{, if nodes } i,j \text{ are visible} \\ 0 & \text{, otherwise} \end{cases} \qquad (2)$$

where the nodes *i*, *j* are denoted as visible, if they can "see" each other, that is, if the segment *i-j* is found entirely inside the area enclosed by the contour (visibility rule). In Fig. 2a it is clearly shown that the nodes 1, 3 form a visible pair, while the nodes 1,2 and 1,4 cannot "see" each other and violate the above mentioned visibility rule. The Visibility Matrix of a cow shape (Fig. 3a) is depicted in Fig. 3b. Note that the notion of visibility matrix with weights counting distance was utilized for shape description to compute the inner distance in [27] but here this concept is utilized by other means. It is important to mention that the visibility matrix A is invariant to translation, scale and rotation, as the visibility rule in not affected by any of these transformations.

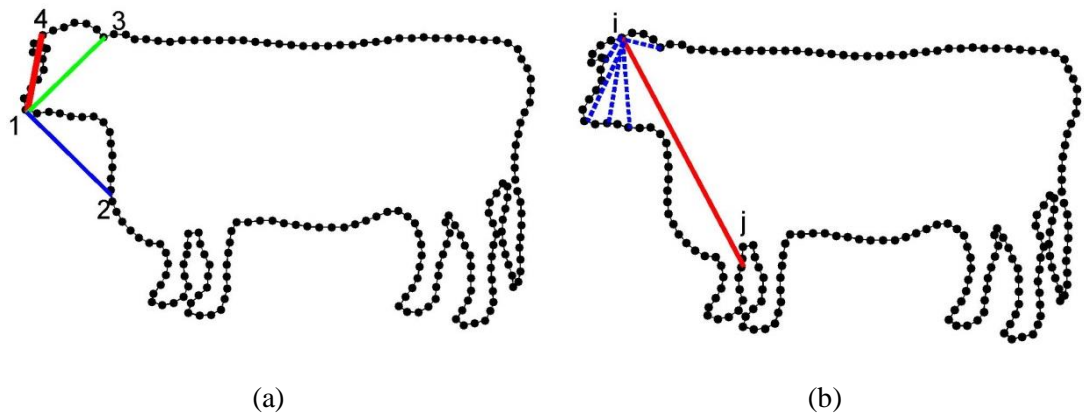

(a)                                                      (b)

**Fig.2** Nodes 1,3 are visible, while nodes 1,4 and 1,2 violate the visibility rule (a). Although nodes $i, j$ are visible they do not belong to the same shape component (b).

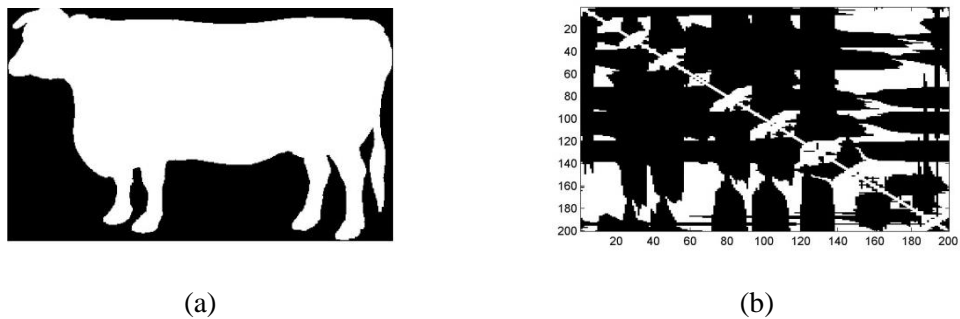

(a)                                                      (b)

**Fig.3** A cow shape for the MPEG7 shape database (a) and the corresponding visibility matrix A, tabulating mutually visible boundary points (b). The black and white representation indicate the 0-1 values of the visibility matrix $A$.

## 3.2 Graph construction

Although the similarity graph $G_{vis}$ that we have constructed directly from the visibility matrix $A$ by connecting all points with positive (one) similarity seems to naturally represent the shape, actually has a shortcoming as it does not capture the local neighborhood. So there emerges the need to reconstruct the graph $G_{vis}$ with additional constraints that models the local neighborhood relationships between the data points.

In our case and in order to facilitate the differentiation of shape parts we define as neighborhood, the number of consecutive contour points. To illustrate this, in Fig. 2b we can see that although nodes $i, j$ form a visible pair, their edge does not contribute in revealing meaningful shape parts, as it links different shape sections. Although nodes $i, j$ can "see" each other, corresponding points are not close (starting from the $i$ boundary point and moving clockwise), thus their possible linking via an edge complicates the shape decomposition problem. Such connections manifest themselves in Fig. 3b as the non-zero values that lie away from the main diagonal. On the other hand, in Fig. 2b nodes linked with node *i* (shown with dotted edges) are found in the same neighborhood and form the cow's head. Therefore the $G_{vis}$ must be updated in order to impose nodes' neighborhood restriction. Specifically, two boundary points are defined as not visible and their corresponding edge is set to zero, if they are placed far from each other in the boundary sequence. Notice that distance is not measured according to the edge length but is calculated as the number *n* of the in-between points when moving along the contour [7]. The edge length information is not utilized in this work as it was found to be less informative towards our objective. It should be also noticed that although there exist many algorithms that transform pair wise similarities of a point set into a graph, like the k-nearest neighbor graphs, the e-neighborhood graphs [5, 6] etc., they are not so appropriate for shape representation.

Consequently, the neighborhood radius *n* -counted in points along the contour- must be defined for each $G_{vis}$ so as to result in the *n*-restricted visibility graph $G_{vis(n)}$ and the corresponding binary visibility matrix $A_n$ $[a_n(i,j)]$. The *n*-restricted visibility matrix $A_n$ (Eq. 4) can be easily obtained by a simple element-wise multiplication of the visibility matrix *A* and the Toeplitz matrix $T_n$ (Eq. 3) defined by the following equations.

$$T_n = toeplitz([I_{n+1} O_{N-(2n+1)} I_n]) \tag{3}$$

where *1k*, *0k* are vectors of length k containing all 1's

and 0's, respectively

$$A_n = A \odot T_n \tag{4}$$

where the $\odot$ denotes element wise multiplication.

In Fig. 4(a-d) we can see $A_n$ matrices for different values of the radius *n*. It is obvious that this parameter determines the neighborhood of each graph and influences the number and the size of the groups that seem to emerge in the main diagonal. A simple, yet not optimal, way to estimate the radius *n* is to calculate the sequence *s* of the off-diagonal summations of matrix *A* for different values of the parameter *n* (Eq. 5) and then find the minima positions. The corresponding values of these extrema are regarded as potential values of the radius-*n*.

$$s(n) = \sum_{\substack{1 \leq i \leq N \\ i \leq j \leq N}} A(i,j), \quad n = 1,2,\ldots,\frac{N}{2} \tag{5}$$

$$s.t. |i - j| = n \text{ or } N - |i - j| = n$$

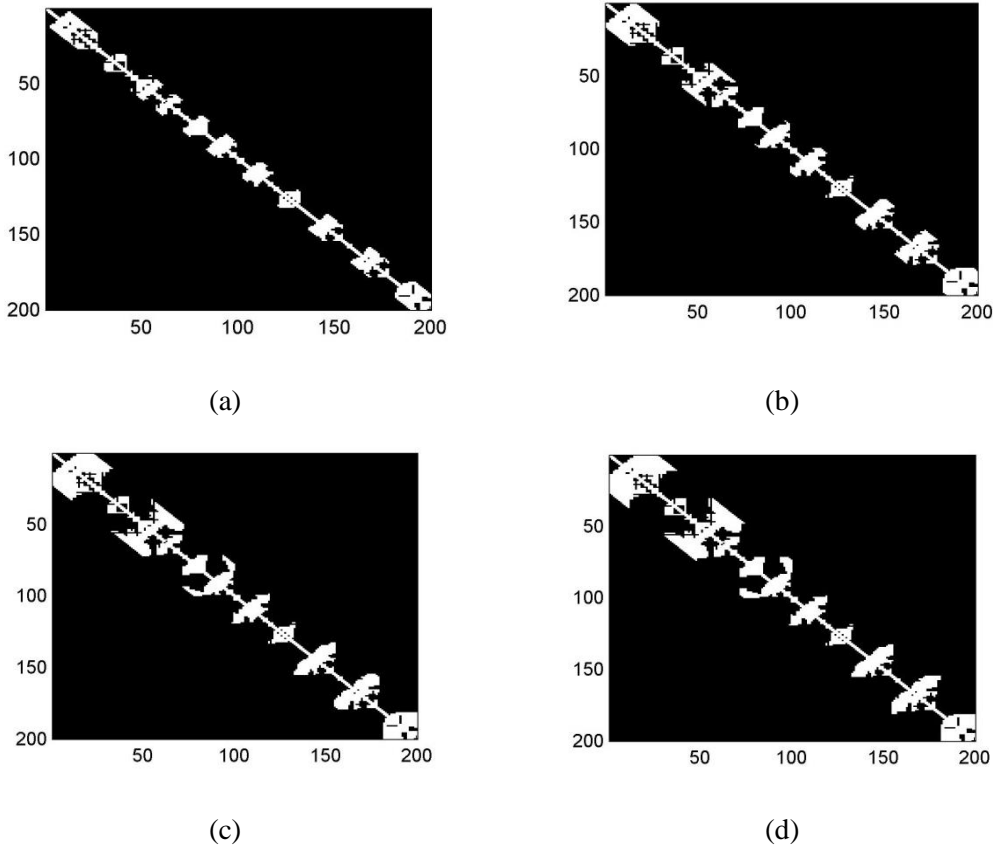

**Fig.4** *n*- restricted visibility matrices $A_n$ for the cow shape for *n* =11, 17, 23 and 27 neighbors respectively (a-d). The value of the parameter *n* affects the number and the size of the groups formed along the main diagonal.

### 3.3 The diffusion matrix

Although the above constructed *n*-restricted visibility graph $G_{vis(n)}$ seems to efficiently capture the local neighborhood of the shape, there are still nodes that should be linked, even though they violate the already mentioned visibility rule. For example in Fig. 5 it is apparent that all nodes *k*, *m*, *i* and *j* belong to the same shape part (i.e. the cow's head). However the nodes *i*, *j* violate the visibility rule, so their corresponding edge is set to zero. Therefore, there arises the need for relaxing these rules. A more insightful view of Fig. 5 reveals that although these nodes do not "see" each other, they are both visible from nodes *k* and *m*. So there exists a route between them through points *m* or *k*. Alternatively, the nodes *i* and *j* are connected by a path of length-two, assuming that each edge is a path of length-one. A known method to capture all these length-two paths in a graph is by simply taking into account the squared matrix of $A_n$. This operation is also considered as a 2 step diffusion process [28]. In general, let *P* be an adjacency matrix of a graph *G* having points $v_1, v_2, ..., v_n$ and *k* be any positive integer. Then the number of distinct paths in *k* steps among points is equal to the corresponding element of matrix $P^k$. In our case we set the value of the integer *k* to 2, as we are interested in diffusing the similarity information into nodes that are not directly visible but have a path of length-two linking them. To that end, let us define the diffusion matrix $D[d(i,j)]$ (Eq. 6), as follows:

$$D=(A_n)^2 \tag{6}$$

This "two-step indirect" visibility facilitates the connection of nodes that belong to the same group, even though they do not all share direct links. It also complies with the concept of clique graphs where each node is connected to all other nodes of the same clique. In Fig. 6(a-d) some examples of diffusion matrices are depicted, for different values of the parameter *n*. Notice the difference between the respective matrices of Fig. 4(a-d). In the case of diffusion matrices it is obvious that blocks seem to be created in the main diagonal.

Clearly, since there are no self-loops, all the elements of the main diagonal are set to zero. It should be noticed that the new graph $G_{diff(n)}$ derived from the diffusion matrix *D* (Eq. 6), is a weighted one with edge weights different than 1'es. This is also depicted in Fig. 6, where diffusion matrices are illustrated with a range of colors, according to their weights, showing the difference to the binary matrices depicted in Fig. 4.

The need for introducing the diffusion matrix *D* arises also by the observation that neighborhood information and diffusion process are complementary operations and accomplish the task of assigning disturbed convex parts -not necessarily visible- to the same cluster. The points *i, j* of Fig. 5 exemplify this observation. These points are not visible but their affinity is reflected in the diffusion matrix where the large number of two-step connections is found and recorded in the *d(i,j)* value.

It should be also noticed that more than two-step diffusion does not contribute to the similarity information of the graph and does not improve segmentation, while complexity is increased. Additionally, in some cases individual parts are grouped together, as a result of separate blocks' merging.

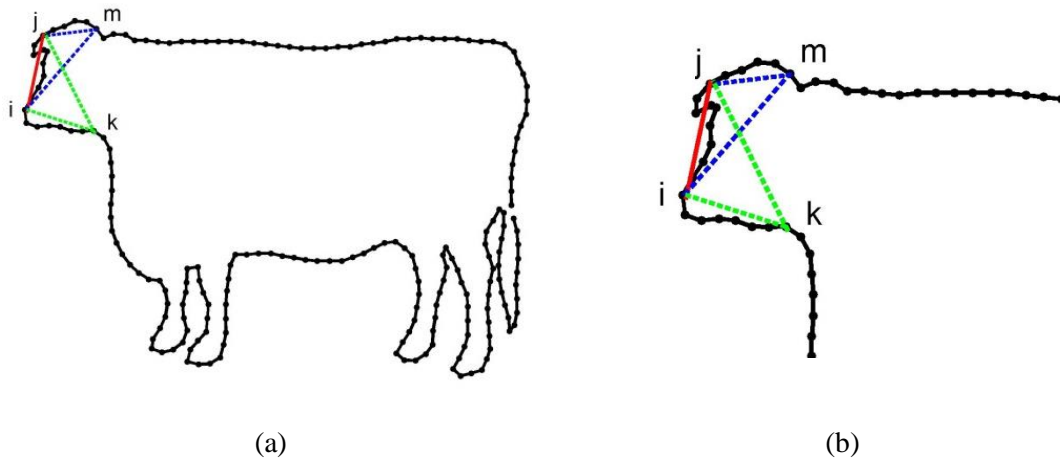

(a) (b)

**Fig.5** Although nodes *i*, *j* violate the visibility rules, their visibility is confirmed by their common visible nodes, i.e. *m* and *k* (a). An enlarged snapshot is depicted for visualization purposes (b).

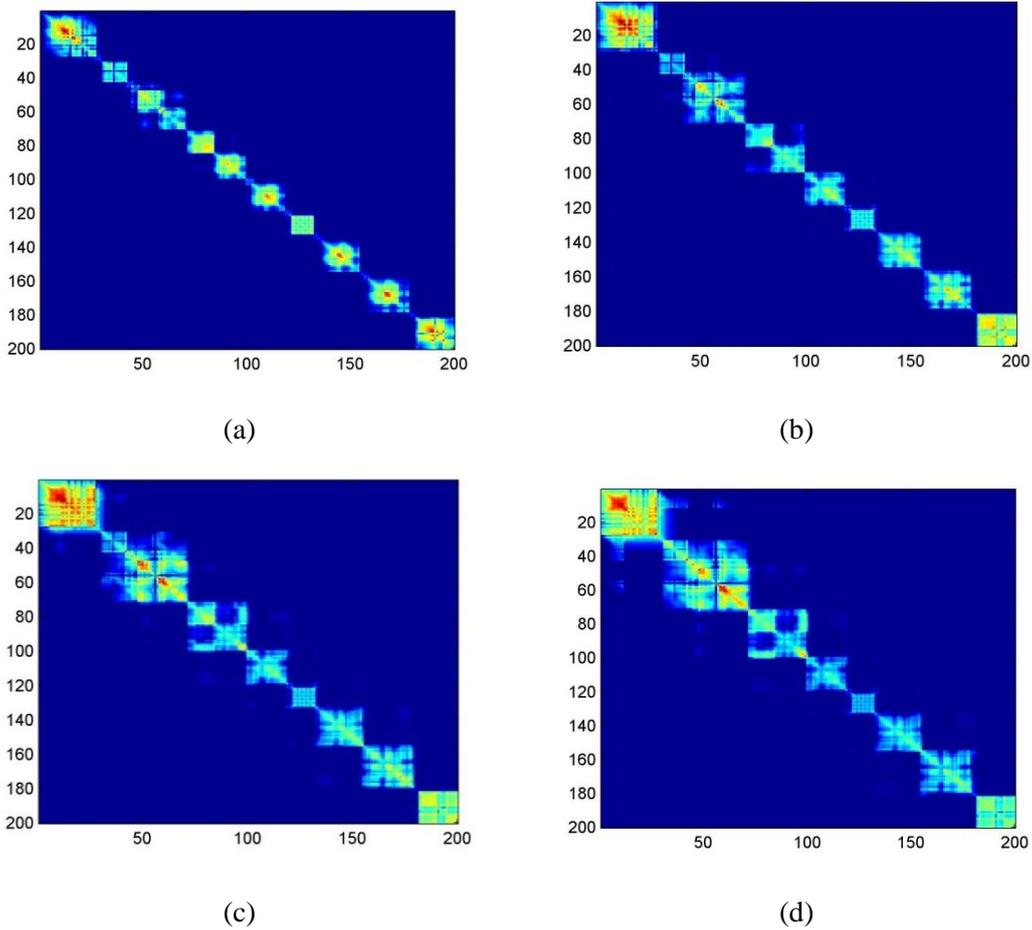

**Fig.6** Diffusion matrices $D$ of the cow shape for different values of the parameter $n$ ($n$=11, 17, 23 and 27 respectively from a to d). Opposed to the respective $A_n$ matrices of Fig. 4, here we can see that blocks seem to reveal themselves along the main diagonal. Note also the colored illustration of $D$ matrices as these are not binary.

### 3.4 The shape decomposition dominant set clustering algorithm

Having transformed the shape into a meaningful graph $G_{diff(n)}$, we now proceed with the final step of our shape decomposition algorithm. The core idea is to separate the shape-nodes into clusters, where each one represents a meaningful shape part. Formally, given a graph representation of a data set, the goal of clustering is to divide the data set into partitions such that the elements assigned to a particular cluster are similar or connected in some predefined sense given that the structure of the graph forms natural clusters. In our case, the diffusion matrix $D$ is appropriately constructed so as the edges are not evenly distributed over the set of vertices, thus avoiding arbitrary clustering results [29].

A common definition of graph cluster is sometimes given by the notion of a clique. Indeed, some authors [13, 30] argue that the maximal clique is the strictest definition of a cluster. Unfortunately, in cases of weighted graphs this concept is not defined. Extension of the maximal clique notion to weighted-graphs is introduced by [14] and is known as the Dominant Set Clustering method. In this approach, as will be shown below, a measure of cluster cohesiveness and a measure of node participation to each group are provided.

### 3.4.1 Problem Formulation

Having defined the similarity graph, our shape segmentation task has converted to a clustering challenge. For graph node set this is equivalent to selecting the group of nodes with large weight values for edges within the cluster and small weight values for edges connecting different clusters. In order to formulate this problem, it is useful to define a participation vector $x$ of length equal to the boundary points $N$, with each entry denoting the degree of the corresponding node participation to a certain cluster. If an entry has small value, then the corresponding node is weakly associated with the cluster. On the other hand, large values of vector $x$ denote strong participation of the respective node to the cluster. Complete absence of node participation is represented by a zero entry in the $x$ vector.

Therefore, a natural way to define the cohesiveness of a cluster can be obtained by the following quadratic expression:

$$J(x)=x^T P x \tag{7}$$

where $P$ is any $NxN$ similarity matrix and $x$ is the participation vector of length $N$.

So the problem is formulated as finding the vector x that maximizes the objective function $J(x)$ (Eq. 7).

Notice that the above mentioned quadratic program was first analyzed by Motzkin and Strauss [31], limited to the case of unweighted binary graphs. Solution of this problem coincides with finding the maximal clique. However, in the dominant set algorithm [14] problem formulation is extended to find the participation vector $x$ and therefore cluster cohesiveness in cases of $P$ being a weighted matrix. In the present method we use the diffusion matrix $D$, with the view of exploiting the important information provided by the presence of distinct paths. Consequently, in order to extract the most cohesive cluster the following standard quadratic program yields:

$$\text{Maximize } J(x)= x^T D x, \text{ s.to} \tag{8}$$

$$x=\{x \in \Re^N\} : x \geq 0 \text{ and } \sum_{i=1}^{N} x_i = 1$$

In Fig. 7 an example is shown, for a simple shape of 200 nodes. The shape (Fig. 7a) is composed of two distinct clusters (depicted in red circles and green stars, respectively). The dominant sets algorithm is applied, using different similarity matrices each time, and the participation vectors of each situation are depicted in Fig. 7b, following the same marking and color rule as that of the clusters. In Fig. 7b where the unconstrained visibility matrix is used, we can observe that the participation vector follows an almost uniform trend at each cluster, which means that each node participates with almost the same strength at the corresponding cluster. Additionally the cohesiveness value is the same for each cluster. In Fig. 7(c,d) where the diffusion matrix $D$ is used with different neighborhood defined by the parameter $n$, the contribution of the node participation vector changes: nodes in the beginning or in the end of the big red circle have restricted visibility to nodes of the same group. However, this behavior is not observed for the small green cluster, where the value of the parameter *n* is greater than the number of participants in the group.

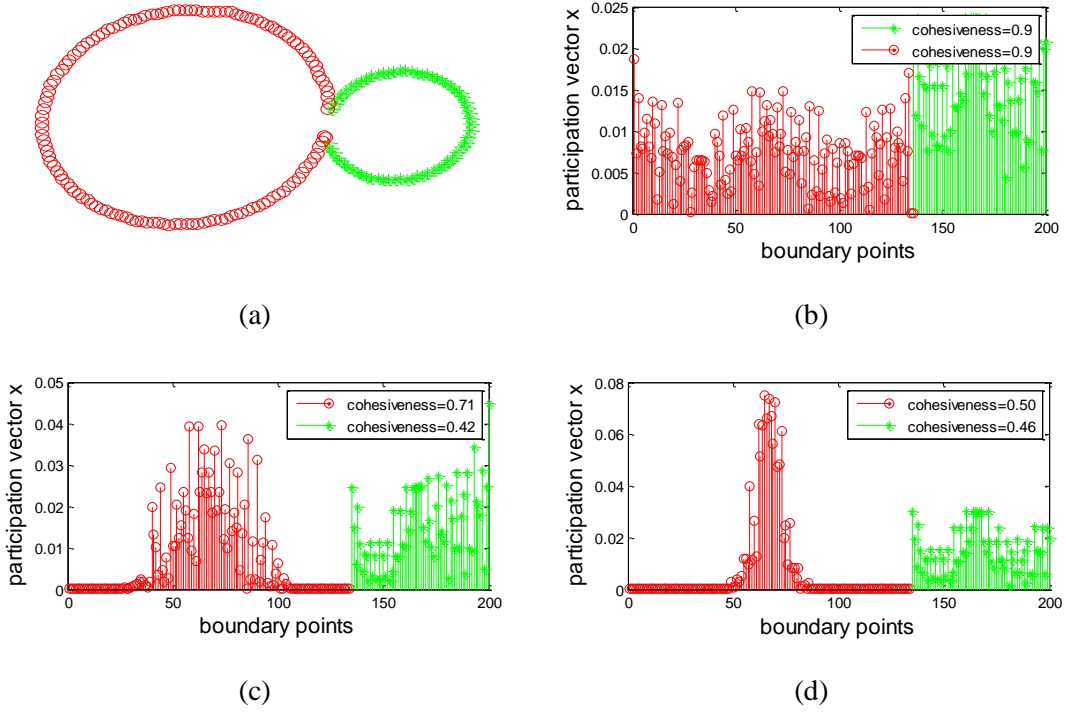

(a)

(b)

(c)

(d)

**Fig.7** An example of the dominant sets method of a simple shape. Distribution of participation vectors using different similarity matrices are shown (a-d). A simple shape of 200 nodes with two obvious clusters (a), dominant set using the visibility matrix (b), dominant set using the diffusion matrix (*n*=96) (c), dominant set using the diffusion matrix (*n*=66) (d).

The dominant sets approach is well suited to our method and with certain modifications we adopt it in the clustering stage of our algorithm. This approach is conceptually similar to the work of [32] where it is stated that the maximum eigenvalue of the adjacency matrix stands for the cluster cohesiveness and the values of the respective eigenvector denote the degree of nodes' participation to the cluster. In other words, the spectrum of the graph provides a clustering of the nodes. However, *x* values can become negative, which has no obvious physical meaning or explanation. Thus, the previously described objective function is not appropriate in our case in the absence of any constraints. Instead in our case, each entry of the vector *x* expresses the corresponding participation into a certain cluster, so the above mentioned constraints (see Eq. 8) are naturally imposed.

A way to find a solution to the problem stated in Eq. 8, was given by the well-known replicator-dynamics [33]. According to that approach the participation vector can be computed through a simple recursive process, provided by the following model (Eq. 9), which we used in our simulations:

$$x_i(t+1) = x_i(t)\frac{(Dx)_i}{x(t)^T Dx(t)}, \qquad i = 1:N \qquad (9)$$

where $x_i(t)$ denotes the participation vector *x* during the *t*-th iteration of the process.

After a fixed number of iterations - i.e. when current iteration does not meliorate more that a tolerance-, the support of *x* is computed providing the set of nodes participating (excluding the zero values one) in the dominant cluster [14].

Regarding the shape decomposition case, an extra constraint should be imposed so as to take into consideration that the cluster nodes (i.e. contour points) constituting a shape part have to be sequential. It should be noticed that the recursive process of Eq. 9 does not guarantee that the high values of the participation vector *x* (i.e. corresponding nodes to be extracted) are in a sequential order. So, a modification of the algorithm is essential.

Each time a dominant set is extracted a potential shape part emerges, which has to be checked whether its constituting nodes are consecutive. If so, then this set of nodes is one of the potential final clusters. On the other hand, in cases that violate the node ordering rule we proceed by selecting the biggest-length sequence of consecutive nodes. The rest of the nodes, which nonetheless may correspond to high participation values in the vector *x*, are neglected and they become potential nodes of any following group. Each time a shape cluster is extracted, the size of the original diffusion matrix *D* is reduced, until the stop criterion is reached (see next Section). The above procedure is repeated until all *N* boundary points are clustered to *m* disjoint clusters:

$V_i$, $i=1:m$ (10)

where $V = V_1 \cup V_2 \cup ... V_\ell ... \cup V_m$ is the set of nodes, $V_i \cap V_j = \{\}$ and $i \neq j$.

The algorithm which describes the extraction of the *m* disjoint dominant sets is summarized in Table I. All entries of the participation vector *x* are initialized with the value 1/*N*, where *N* is the number of boundary points. Classes with less than four participants are ignored.

**Table I**. The iterative dominant set algorithm

Input: Visibility Matrix A

Output: The set of all m dominant disjoint clusters

1. Use Eq.4 to construct the n-restricted visibility matrix $A_n$.

2. Use Eq.6 to construct the diffusion matrix D.

3. Initialize the participation vector x.

4. Use Eq. 9 to find a local solution to the problem of Eq.8.

5. Apply the sequential constraint and find x.post

6. Extract the ith cluster, i=1:m.

7. Update D according to x values.

8. If i<m go to step 3. Else return.

### 3.4.2 Determining the number of clusters: random graphs

The above mentioned iterative procedure (Table I) terminates when all nodes are assigned to a cluster. However, we are interested in capturing the shape parts which means that not all clusters are essential, as there exist shape nodes that do not belong to a specific part. Therefore, we are interested in discovering a set of $k$ dominant clusters ($k \leq m$), where each cluster includes the group of sequential nodes that correspond to a specific shape part. The number $k$ of the shape segments is not necessarily the same as the number of the $m$ disjoint clusters (Eq. 10). For that reason, the algorithm described in the previous subsection is successively applied until the desired number of the $k$ clusters is extracted. Notice that in cases where the number $k$ of the finally extracted clusters is less than $m$, it means that some nodes of the original set $V$ remain unassigned to a cluster. In fact, there exists an optimal threshold value that determines the meaningful classes, which will be the focus of the section that follows.

In order to determine the set of $k$ clusters that we are going to keep, the notion of cohesiveness is used. The main idea is that groups with great cohesiveness, are most important than the rest, and naturally this is the rule of selecting the correct clusters.

Therefore, a threshold value should be set, determining which clusters are of greater importance compared to others, of smaller cohesiveness value. Despite the fact that this seems to be a quite reasonable and simple notion, there arise two issues that need to be addressed: first, the exact value of the threshold which defines the number of teams that are going to be kept and second the need of adopting a different threshold value at each shape-graph.

In order to successfully tackle these problems, we employ a graph randomization technique, and particularly we consider null-hypothesis graphs, which have simple random or ordered topologies but preserve basic characteristics of the original network. However a standard process for generating uniform random graphs [34] where the edges are distributed uniformly and the degree distribution is Poissonian, is not appropriate for the present method.

Particularly, we will employ an algorithm that randomizes the graph, while preserving the degree distribution. The degree is very important for the graph topology that is implied by the diffusion matrix $D$. From the construction of matrix $D$ one can easily understand that the degree distribution is changed in an efficient manner that facilitates revealing the shape parts, while new edges are created. Most of the nodes that belong to a shape part have a higher degree than those that are left unassigned and additionally it is observed that in many cases, same degree nodes belong to the same cluster. Furthermore, in the corresponding weighted graph an edge value is augmented each time there exist a node which is mutually visible by the endpoints of that edge. Note that in cases of originally visible nodes, an additional 1-value is added to the corresponding edge. All these mean that the edge - strength plays an important role to the clustering procedure, too. However, because the degree is quite high in the most significant clusters, we can guide the output by favoring a degree-preserving rewiring algorithm, which is additionally less computational expensive.

The method of random graphs described above constitutes the core of the second part of our algorithm presented in Table II which produces the correct final clusters $k$ ($k \leq m$) that define the shape parts. In the first step we construct a randomized graph [35], originating from the diffusion matrix $D$. The graph randomization algorithm that preserves the degree distribution is used next. Notice that matrix $D$ is connected because there always exist a path from one node to another (neighbors with $n=1$ are visible by default). Using that graph, we apply the dominant set algorithm, as described in previous Section. Then, we find the clusters and their respective cohesiveness. We repeat the above procedure for a set of random graphs. The mean value of all the cohesiveness' values plus two standard deviations is the threshold value for the original matrix $D$. The cohesiveness value $J(x)$ is calculated in this paper by assigning at each entry of the participation vector x the value $1/N_c$, where $N_c$ is the number of nodes participating at the specific dominant set. An outline of the algorithm that determines the number $k$ of final clusters is presented in Table II.

---

**Table II.** The algorithm for final k clusters determination

Input: Diffusion matrix D

Output: The set of k final dominant clusters, k≤m

1. For i=1:NumberOf RandomGraphs

  2. Use the graph randomization algorithm to construct a random graph from D.

  3. Run the iterative dominant set algorithm using the random graph.

  4. Use Eq.7 to calculate cohesiveness for each cluster.

5. End

6. Threshold=mean(Cohesiveness)+2std(Cohesiveness).

7. Run the iterative dominant set algorithm (Table I) using the D matrix.

8. Use Eq.7 to calculate cohesiveness for all m clusters.

9. Use Threshold value and keep k clusters, k≤m.

---

## 4. Experimental Results

Experiments of the proposed method on three shape databases are conducted, as well as comparative results using other shape decomposition methods. The experimental setup is the following: all boundary shapes are uniformly downsampled into 200 or 250 points, the number of random graphs generated each time to select the appropriate threshold is set to 250.

### 4.1 2D shape decomposition

Our method is applied to the MPEG-7 CE-shape-1 part B [36] shape database and the Kimia99 [37] shape database. The first database consists of 1400 shapes, classified uniformly into 70 categories and the Kimia's database consists of 99 shapes classified uniformly into 9 categories.

In Fig.8 a sample of decomposed shapes of the MPEG7 database, using the proposed method is depicted. As we can observe, the method seems to be insensitive to the number as well as the complexity of the shape components. In addition, shapes with bendable parts are included in order to show the effectiveness of the proposed method even when shapes are depicted in different poses. It is evident that for most cases decomposition results are meaningful. Specifically, the dog's main components are successfully found (head, tail, legs), the mouse is also segmented into a set of natural parts (ears, tail, hands, legs) etc. Poor performance of the algorithm can be found in few cases (i.e. the elephant's back) but can be corrected by using an appropriate post-processing procedure which is described in the next paragraph.

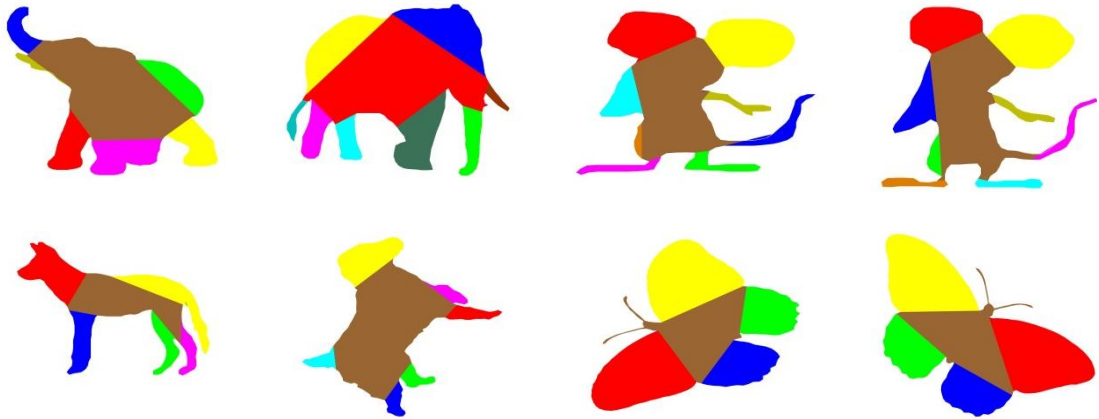

**Fig.8** A sample of decomposed shapes of the MPEG7 database, using the proposed method.

Furthermore, in Fig. 9 a sample of results from the Kimia database is shown and is indicated that the proposed method results in good partitioning, even when some occlusions or distortions appear.

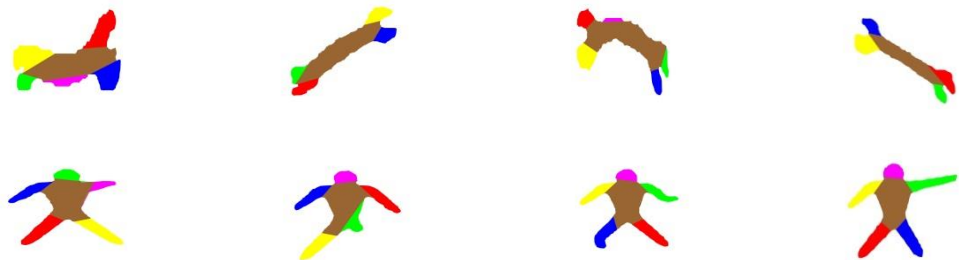

**Fig.9** A sample of Kimia database. Despite several occlusions and distortions, the algorithm succedded in decomposing the shapes into their parts.

To further demonstrate the effectiveness of our decomposition method we present some comparative results, in Fig.10. In particular, our method is compared to [38], [23], [24] and [39]. In the first column of this figure, human decomposition results are given. Human perception is the undisputed evaluator of the produced results. In detail, for each one of the categories of Fig.10 humans were asked to manually separate the shape parts and this column will serve as the ground truth of our comparison results. The results from this experiment where borrowed from [39]. Also, we should mention that for the results of method [39] shown in the fifth column, the "straightening" process they propose, is not included. For visualization purposes, the results of our method are registered into a rectangle.

As we can see from Fig.10 the bone, the margaret and the bream shapes are in full accordance with the human perception results, given in the first column. Notice also that the camel and the octopus follow the same attitude and none of the other methods end to same results (for example the other methods fail to capture the octopus's legs as separated components). Continuing, in the bug and the chicken shapes, our method captures the legs as whole components, while in the other comparative results the legs are over-segmented. The fork shape is almost identical to the perceptual-based decomposition, except from some points that are left unassigned. Moreover, in the cow shape, although an extra part is found, all basic components are captured – such as the legs, tail and head.

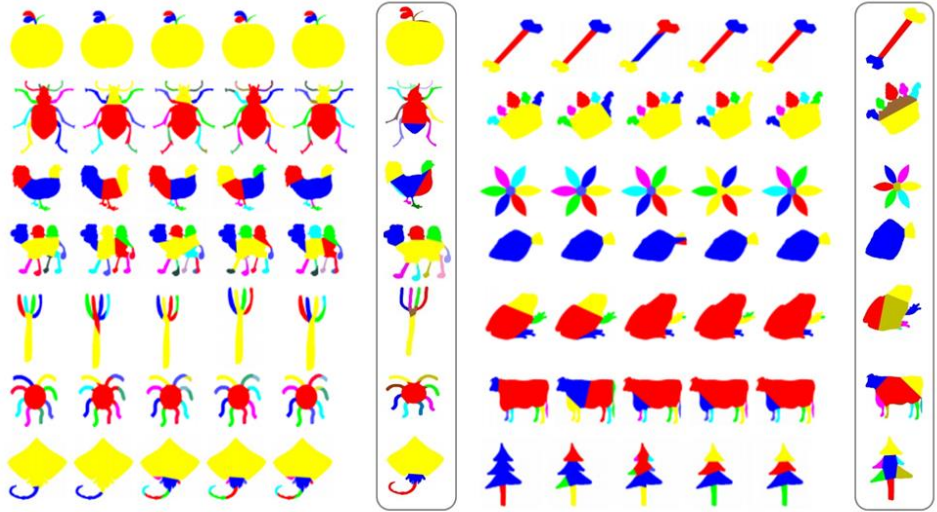

**Fig.10** Examples of decomposition results for 14 categories of the MPEG7 shape database. Human decomposition results (1st column), [38] results (2nd column), [23] results (3rd column), [24] results (4th column), [39] results (5th column) and our decomposition results, registered into a rectangle (last column).

- **Improving the results- post processing**

In this subsection, an optional step for post processing the outputs is described. This post processing refers to a final step which effectively utilizes the unassigned nodes produced in the randomization stage. The concept is to check for weak shape parts by counting their "affinity" to the unassigned nodes. This is easily quantified by computing the aggregate visibility between nodes of each cluster $C=1,2,\ldots,k$ and nodes that are left unassigned which are essentially points that do not belong to a specific or dominant shape part (i.e. the $m-k$ clusters that the algorithm intentionally discarded). The cluster that shares the greatest visibility with the unassigned nodes is mostly unlikely to form a meaningful shape part and is removed from the set of the $k$ clusters and merged with that set. To that end let us define the quantity $q$ for each cluster:

$$q(C) = \sum_{i,j=1}^{|V_c \cup V_u|} a_{ij}^{V_c \cup V_u} - \left( \sum_{i,j=1}^{|V_c|} a_{ij}^{|V_c|} + \sum_{i,j=1}^{|V_u|} a_{ij}^{V_u} \right) \quad , C=1,2,\ldots,k \quad (11)$$

where $A^{V_\ell}$ denotes the subpart of the visibility matrix $A$ concerning the nodes of set $V_\ell$,

$V_u = \{V_{k+1} \cup ... \cup V_m\}$ defines the set of nodes that are left unassigned

$V_C$ defines the set of nodes that belong to cluster C

and |B| denotes the cardinality of set B.

The cluster $C$ that maximizes the above formulated quantity $q$ is merged with the set of unassigned points $V_u$. Application of the above simple method to our resulting clustering improved the shape decomposition output. Some examples are depicted in Fig. 11 where redundant- meaningless parts are eliminated.

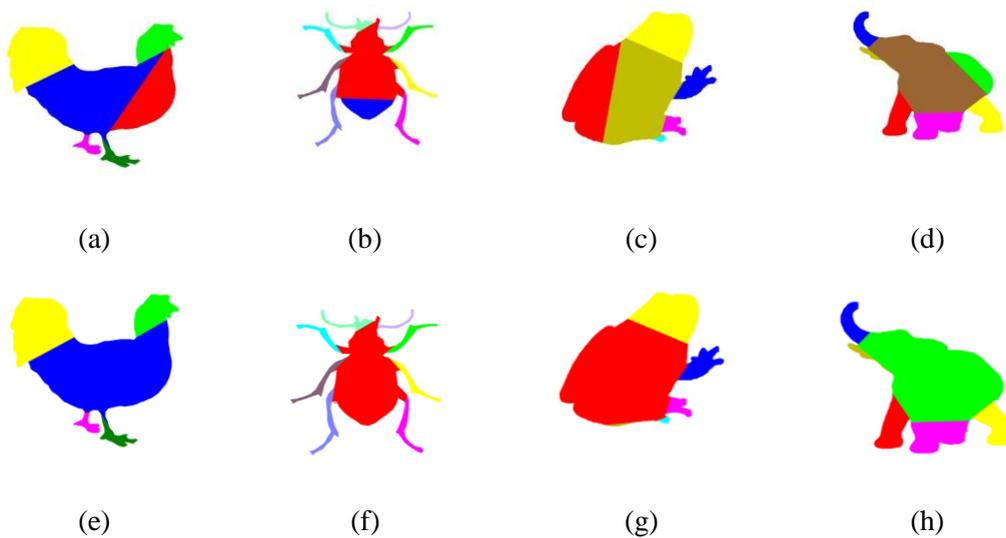

(a)　　　　　(b)　　　　　(c)　　　　　(d)

(e)　　　　　(f)　　　　　(g)　　　　　(h)

**Fig.11** Examples of decomposition results before the post-processing procedure (a-d) and after the post-processing procedure (e-h). Note that redundant parts are ignored.

### 4.2 Hand gesture decomposition

In this subsection we will present sample results from the hand gesture database provided by [40]. For more information concerning the necessity of handling hand gesture images, please refer to [41]. A sample of this database images is depicted in Fig. 12, where we can see two different gesture classes (i.e. the V-shape and the spread). It is essential to notice that these images depict real circumstances –unlike the binary shapes of the previously shown databases- and additionally there exist different illumination conditions which hinder the contour extraction procedure. All of the above, lay the decomposition problem more challenging. Results using the introduced decomposition method are given in Fig. 13. Despite the fact that the images are captured under various illumination, orientation and scale conditions, the decomposition results are very satisfactory. For this database, the threshold value in order to obtain the appropriate number of classes was set to mean cohesiveness value plus four standard deviations.

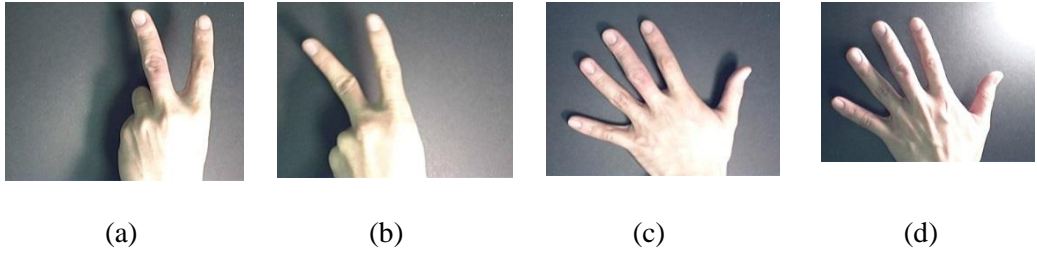

(a) (b) (c) (d)

**Fig.12** A sample of the hand gesture database. The V-shape (a,b) and the spread (c,d) are shown. Notice also the different illumination conditions.

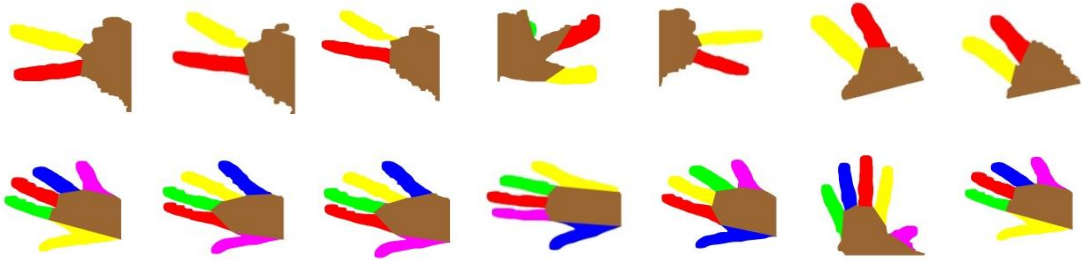

**Fig.13** Hand gesture recognition using the proposed method, on the V-shape and the spread. Notice the satisfactory results, despite the different illumination conditions of the original images.

## 4.3 Quantitative Results

In addition to the qualitative results, in this section we will provide some quantitative comparisons which aim to show that our method results in a good clustering, compared to some other clustering results. There exist some desirable properties that a graph clustering should have. It is generally agreed that a subset of vertices forms a good cluster if the induced subgraph is dense, but there are relatively few connections from the included vertices to vertices in the rest of the graph, keeping in mind the overall graph's density [29]. In order to measure all the above, the following formulas are employed [29]. The density of a graph $G(V,E)$, is defined as the ratio of the number of edges present to the maximum possible (Eq. 11). The internal cluster density (Eq. 12) counts the density of each subgraph, i.e. of each output cluster and in order to calculate the internal density of the whole clustering, an averaging procedure takes place (Eq.13) while the external density tries to compute the cut sizes of all clusters (Eq. 14).

$$\delta(G) = \frac{|E|}{\binom{|V|}{2}},\tag{12}$$

where |.| denotes the cardinality of the corresponding set.

$$\delta_{int}(C) = \frac{|\{(v,u)\ s.t.v\in C, u\in C\}|}{|C|(|C|-1)},$$

(13)

where $C$ denotes one of the $k$ clusters $C_1....C_k$ and $v, u$ denote the nodes.

The internal density of a given clustering of a graph $G$ into $k$ clusters $C_1, C_2, \ldots, C_k$ is the average of the internal cluster densities of the included clusters:

$$\delta_{int}(G\backslash C_{1,...,}C_k) = \frac{1}{k}\sum_{i=1}^{k} \delta_{int}(C_i) \qquad (14)$$

$$\delta_{ext}(G\backslash C_{1,...,}C_k) = \frac{|\{(v,u)\ s.t.\ v \in C_i, u \in C_j, i \neq j\}|}{|V|(|V|-1) - \sum_{\ell=1}^{k}(|C_\ell|(|C_\ell|-1))} \qquad (15)$$

For a good clustering, the internal graph density $\delta_{int}(G\backslash C_{1,...,}C_k)$ should be notably higher than the graph density, while the external graph density $\delta_{ext}(G\backslash C_{1,...,}C_k)$ should be lower than the graph density $\delta(G)$.

In an attempt to quantify the clustering performance of the proposed method using the above mentioned formulas, we intentionally produced some individual decompositions that fail to capture appropriately the meaningful parts of a shape. In Fig. 14 we can see the cow shape of the MPEG7 database in such different decomposition results, i.e. extending the head cluster, merging two clusters or ignoring one cluster and for each case the indices of Eq. 11-14 were calculated. Notice that some of these individual results can be found in the compared methods depicted in Fig. 10. Observe that the graph density has the same value (0.3089) for all cow decompositions due to the visibility graph which is the same in all cases. As already mentioned, in order to have a good clustering the following inequality should hold: $\delta_{ext} < \delta << \delta_{int}$ and it is obvious that our proposed method satisfies better both sides of this inequality.

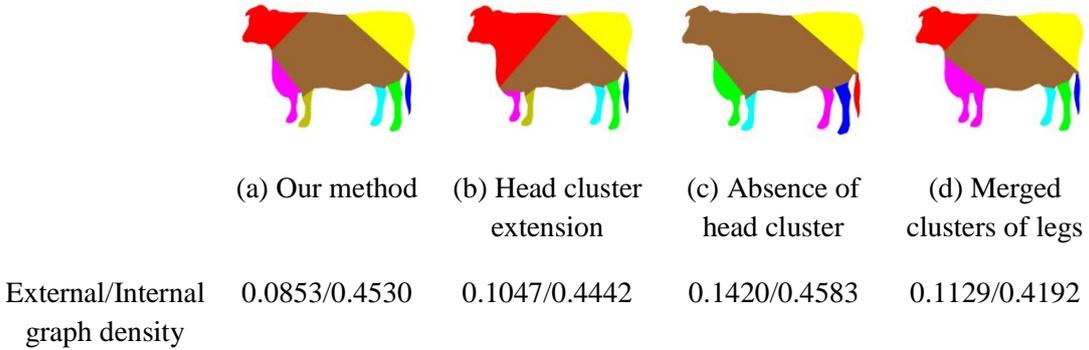

| | (a) Our method | (b) Head cluster extension | (c) Absence of head cluster | (d) Merged clusters of legs |
|---|---|---|---|---|
| External/Internal graph density | 0.0853/0.4530 | 0.1047/0.4442 | 0.1420/0.4583 | 0.1129/0.4192 |

**Fig.14** Clustering indices for different shape decomposition results. Our method satisfies better both sides of this inequality $\delta_{ext} < \delta << \delta_{int}$.

Moreover, quantitative results are provided in comparison with the method presented in [9]. The reason we choose this method is that it is also a graph-based one that additionally constructs a visibility graph. Notice that the resemblance of these methods is restricted only in the first step, i.e. the visibility matrix. For the sake of a fair comparison, the number of clusters is set to the same value for both methods. In Table III results of our method against the method of [9] are presented. Regarding the inequality that should hold in order to have a good clustering it is obvious that the proposed dominant sets based method outperforms that of [9]. This is probably due to the fact that in the latter one the neighborhood is not captured and the visibility graph contains edges that connect nodes from different shape parts.

| Table III. Comparative quantitative measurements of our method to [9]. | | | | |
|---|---|---|---|---|
| MPEG7 shape | Our method | | Method of [9] | |
| | $\delta_{ext} < \delta << \delta_{int}$ | Q | $\delta_{ext} < \delta << \delta_{int}$ | Q |
| beetle | 0.0560<0.1860<<0.4392 | 0.7415 | 0.3591<0.2018<<0.3681 | 0.5017 |
| camel | 0.0707<0.3151<<0.4828 | 0.6102 | 0.1718<0.3318<<0.3705 | 0.4016 |
| octopus | 0.0700<0.3325<<0.4874 | 0.5858 | 0.1618<0.2878 <0.3632 | 0.4249 |
| cow | 0.0853<0.3089<<0.4530 | 0.6775 | 0.2267<0.5588 <0.4007 | 0.1878 |

Another well-known quality evaluation metric for graph clustering is the modularity [42], which actually states that a good cluster should have a larger than expected number of internal edges and a smaller than expected number of inter-cluster edges when compared to a random graph with similar characteristics. Modularity formulation is given by the following equations:

$$Q = \sum_{i=1}^{k}(e_{ii} - a_i^2), \quad (16)$$

where

$$e_{ii} = \frac{|\{(v,u) s.t. v \in C_i, u \in C_i\}|}{|E|} \quad (17)$$

$$a_i^2 = \frac{|\{(v,u) s.t. u \in C_i\}|}{|E|} \quad (18)$$

Values approaching $Q=1$, which is the maximum, indicate good clusterings, i.e. with strong cluster structure. In Table III comparative results are provided and it is clear that the proposed method reaches values of greater modularity instead of the method introduced in [9].

To continue, a measure of the quality of the clustering could rely on the number of output parts [24]. Although resulting in a minimum number of parts is not always in compliance with the human's decomposition, it is a simple way to measure the degree of parts' redundancy. Therefore, this measure is more appropriate in cases of shapes with bendable parts, which are commonly over segmented. As introduced in [24], the reduction rate score (↓) – compared to a method- is defined as:

method ↓= (#method − #ourMethod)/# method where method can be any method we are interested to compare with and the symbol # denotes the number of parts. In terms of the above mentioned reduction rate score, in Table IV our method is compared to the methods of [38], [23], [24] and [39]. The shapes we chose for comparison purposes include - but not limited to - bendable parts, as these parts are commonly over-segmented, which is obviously wrong. A robust method should be able to output these parts as a whole and do not proceed in further partitioning. As we can see, our method results in a small number of parts, compared to mostly any one of the other well-known techniques, which proves that it can deal successfully with shapes of bendable parts. Although this measure is well suited for such shapes, the results for rigid shape-parts are also most of times in agreement with the ground truth, which is also depicted in the Table IV.

At that point however, we have to mention that a method that produces the least number of parts is not always consistent with the human perception. For example in the octopus case, the method of [38] produces 8 parts, whereas it is obvious (see Fig.10) that 9 parts is in fully accordance with the human's decomposition. The same holds with the margaret-like shape.

| Table IV. The average reduction rate score (%) of our method compared to the methods of [38], [23], [24] and [39]. | | | | |
|---|---|---|---|---|
| MPEG7 shape | [38] | [23] | [24] | [39] |
| beetle | 0 | 0.25 | 0.36 | 0.31 |
| ray | -0.5 | 0.57 | 0.57 | 0.57 |
| camel | 0 | 0.25 | 0.1 | 0.19 |
| octopus | -0.125 | 0.1 | 0 | 0.19 |
| apple | -0.5 | 0.25 | 0.25 | 0.25 |
| elephant | 0.14 | 0.25 | 0 | 0.14 |
| margaret | 0 | 0 | 0 | -0.75 |
| chicken | -0.7 | 0.3 | 0.17 | 0.17 |
| beam | 0 | 0 | 0.5 | 0 |

Finally, another well-known measure to evaluate part segmentation is the Rand Index (RI) [43]. A modified version is used here, where only contour points are taken into account, as in the 2D shape decomposition task it is the contour points that actually define the segments. In addition, in that way, we do not favor parts that occupy great area over the shape and therefore, small,

usually bendable parts are not discarded. The RI for two segmentations, $S_1$, $S_2$ is defined as follows:

$$RI(S_1, S_2) = \binom{2}{N}^{-1} \sum_{i,j,i<j} [C_{i,j}P_{i,j} + (1-C_{i,j})(1-P_{i,j})] \qquad (19)$$

,where $N$ denotes the number of contour points and is set to *200* for the Kimia database and

$C$, $P$ are the $N$x$N$ binary matrices which elements are set to one if the nodes $i$, $j$ belong to the same segment, else they are set to zero.

Using the above-formulated measure, we conducted two experiments which we discuss in the following paragraphs. In both experiments, all methods are compared to the human segmentations. Each shape is evaluated using the corresponding human segmentations. In addition, humans' decompositions are also compared which is regarded as the baseline.

For the first one, we used a subset of the Kimia99 database, using *50* randomly selected shapes. For ground truth, we asked *10* humans to decompose the same shapes according to their judgment. In the Table V, we present the average RI for all the shapes, accompanied by the standard deviations (STD). From Table V, we see that the average RI value of the proposed method almost the same with that of human, which means that the introduced method provides segmentations close to the aggregated human ones and as we can see, while the STD value of the proposed method is less than that of the human.

The second experiment aims to compare the method with the techniques proposed in [22], [24] and [44]. The shapes used for this experiment, as well as the corresponding human segmentations, are available in [44]. Notice that in our case, only shapes with a single closed contour were included. The RI values are shown in Table VI and as we can see the introduced method achieves higher score than the methods [24], [22] and a little lower than that of [44] and human baseline. This evaluation confirms that the resulting decompositions of our method are close to the ground truth and outperform or are close to other widely-known methods.

| Table V: Mean RI values comparing the proposed method with the ground truth. Human vs human comparison is regarded as the baseline. The corresponding STDs are also provided. | | |
|---|---|---|
| method | proposed method | human |
| RI | 0.7661 | 0.7870 |
| STD | 0.1739 | 0.2301 |

| human | [44] | [24] | [22] | Proposed method |
|---|---|---|---|---|
| 0.73 | 0.68 | 0.46 | 0.39 | 0.63 |

Table VI: Mean RI values comparing the proposed method with the methods of [44], [24] and [22]. Human vs human comparison is regarded as the baseline.

## 5. Conclusions

In this paper the problem of decomposing a shape into its meaningful parts by adopting a graph-based approach is addressed. In the graph construction stage the shape is initially represented by a binary visibility graph in which the constraint of capturing local neighborhood is additionally imposed. Next, the binary graph is transformed into a diffusion (weighted) one, where information is efficiently spread through the graph's nodes. For the graph clustering stage that follows we showed how the notion of cliques and the dominant sets can serve as a solution to the shape decomposition problem. A basic output of this algorithm is the cluster's cohesiveness, which we additionally exploited in order to automatically determine the number of shape segments, appropriately combined with a random network generation procedure. Qualitative and quantitative experimental results confirmed the effectiveness of the proposed method.